\documentclass{article}
\usepackage{ijcai24}

\usepackage{times}
\usepackage{soul}
\usepackage{url}
\usepackage[hidelinks]{hyperref}
\usepackage[utf8]{inputenc}
\usepackage[small]{caption}
\usepackage{graphicx}
\usepackage{amsmath}
\usepackage{amsthm}
\usepackage{booktabs}
\usepackage{algorithm}
\usepackage{algorithmic}
\usepackage[switch]{lineno}
\usepackage{xcolor}
\usepackage{comment}
\usepackage{enumitem}
\usepackage{amssymb}

\usepackage{algorithm}
\usepackage{algorithmic}

\newtheorem{definition}{Definition}

\def\rvb{{\mathbf{b}}}

\def\rvm{{\mathbf{m}}}

\def\rvp{{\mathbf{p}}}

\newcommand{\hv}[1]{\mathbf{#1}}

\newcommand{\basis}{\mathbf{B}}

\newcommand{\domain}{\mathit{D}}

\newcommand{\concept}[1]{\mathit{#1}}

\newcommand{\proto}[1]{\rvp_{\concept{#1}}}

\newcommand{\dime}[1]{\mathrm{dim}(#1)}

\newcommand{\bind}{\circledast}

\newcommand{\real}[1]{\mathbb{R}^{#1}}

\newcommand{\normal}[1]{\hat{#1}}

\newcommand{\vasanth}[1]{\textcolor{blue}{\textbf{[Vasanth]: #1}}}

\title{Analogical Reasoning Within a Conceptual Hyperspace}

 \author{
 Howard Goldowsky
 \and
 Vasanth Sarathy
 \affiliations
 Tufts University\\
 \emails
 \{howard.goldowsky, vasanth.sarathy\}@tufts.edu
 }

\begin{document}

\maketitle

\begin{abstract}
    We propose an approach to analogical inference that marries the neuro-symbolic computational power of complex-sampled hyperdimensional computing (HDC) with Conceptual Spaces Theory (CST), a promising theory of semantic meaning.
    CST sketches, at an abstract level, approaches to analogical inference that go beyond the standard predicate-based structure mapping theories. But it does not describe how such an approach can be operationalized. We propose a concrete HDC-based architecture that computes several types of analogy classified by CST. We present preliminary proof-of-concept experimental results within a toy domain and describe how it can perform category-based and property-based analogical reasoning. 
\end{abstract}

\section{Introduction and Motivation}
\label{sec:introduction}

``Analogies are partial similarities between different situations
that support further inferences.'' \cite{Gentner_1998} The well-known formulation

\begin{equation}\label{eq:analogy}
    \concept{A} : \concept{B} :: \concept{C} : \concept{X}
\end{equation}
represents an analogy. Analogical reasoning entails many key aspects of human cognition and involves several key processes: \textit{retrieval} (given $\concept{C}$, find $\concept{A}$ and $\concept{B}$), 
\textit{mapping} (determine a structural correspondence between $\concept{A}$ and $\concept{B}$ to find $\concept{X}$, by applying the correspondence to $\concept{C}$), and \textit{inference} (using $\concept{A}$ to advance the concept $\concept{C}$).  In this paper, we focus on the task of \textit{mapping} -- more specifically, the task of identifying a relationship between $\concept{A}$ and $\concept{B}$, and then applying the identified relationship to characterize $\concept{X}$. The particular challenge with mapping is that there is often a large number of potential relationships between $\concept{A}$ and $\concept{B}$, and these relationships may themselves be compositional and graded in nature, as well as span the symbolic/sub-symbolic representational divide. Finding the \textit{salient} relationships -- the ones between $\concept{A}$ and $\concept{B}$ relevant to $\concept{C}$ -- is a combinatorially hard problem.  

Approaches to solving the mapping problem have been either connectionist or symbolic, and both types of models attempt to identify structural correspondence (graph isomorphisms) between concepts \cite{Gayler_Levy_2009}. Connectionist approaches such as ACME and DRAMA \cite{Eliasmith2001} are essentially ``localist'' \cite{Page_2000} in nature, which means concept representations/symbols, although connected within networks, remain localized to single nodes. Purely symbolic approaches, on the other hand (such as Structure Mapping Theory \cite{GENTNER1983155,Forbus_2021}), explicitly incorporate the geometric graph structure using a predicate-based representation. Both types of analogy engines remain constrained by their representations. 

Localist-connectionist approaches require a decomposition and sequential symbolic traversal of the source and target structures, substantially increasing their time complexity \cite{Gayler_Levy_2009}. Purely symbolic approaches struggle to compute analogies that need to isolate salient conceptual properties of objects \cite{Gardenfors2023-GRDRWC}. Salience requires a distance metric that just does not exist within a symbol-dominated space.

Neural networks offer more of a ``distributed'' connectionist approach. Unfortunately, however, their concept embeddings do not necessarily possess the structural and compositional aspects with which to perform analogical mapping. Moreover, neural network models fail to generate many types of analogies outside the distributions that characterize their training sets, and they fail to provide the underlying hierarchical structure to support analogical inference \cite{Mitchell,Lewis2024UsingCT}. A cognitive framework with the ability to simultaneously and seamlessly represent these so-far mutually exclusive connectionist and symbolic computational paradigms has been lacking \cite{LIETO20171}.

More recently, researchers have explored hyperdimensional computing (HDC), synonymously known as vector-symbolic architecture (VSA)\footnote{To ensure clarity, we will use the term hyperdimensional computing or HDC in the rest of this paper.} as a paradigm for capturing a number of neurally plausible cognitive phenomena, including analogical mapping \cite{HerscheRPM,Kanerva1997FullyDR,Plate_2003,Gayler_2004,Blouw}. HDC is a representational and inferential paradigm in which data structures can be represented with high-dimensional vectors, thus bridging the symbolic/subsymbolic gap.

The study of analogical inference using HDC is relatively new. Many open questions exist. In \cite{Navneedh_Maudgalya_Olshausen_2020}, for example, the salient aspects of the $A:B$ relationship were given, but it was assumed that the structure underlying $A$, $B$ and $C$ were binary. The question of how to identify more salient aspects of structure and allow for a more nuanced or graded characterization of concepts remains open. 

In this paper, we begin answering this question with a crucial insight: that the HDC framework could operationalize the cognitive science theory of Conceptual Spaces \cite{Gardenfors2000}. For over a quarter century, Conceptual Spaces Theory (CST) has been an attractive guide for research in cognitive science \cite{Douven2019}, neuroscience \cite{BellmundSpatialCodes}, and artificial intelligence \cite{ACS_Zenker}. Its strength lies in its inclusive ability to model how sensory observations flow through an agent's initial connectionist layer of observation processing, through a layer of geometric concept space, and ultimately into a form of symbolic representation often used for reasoning -- thus sharing many properties with HDC.

CST  offers guidance on analogical mapping. Under CST, analogical inference requires a distance metric within a concept space. This requirement implies at least five necessary and simultaneously present capabilities of an agent \cite{VelezGardenforsAnalogyAsSearch}: 
\begin{enumerate}
    \item It must continuously accept sensory observations in real-time and encode these signals into working memory;
    \item It must afford an algebraic calculus over distance metrics within working memory; 
    \item It must afford a logical calculus in working memory; 
    \item It must cross-reference the same concepts during sensory observation, geometric processing, and symbolic processing;
    \item It must provide interaction with long-term memory in an associative capacity.
\end{enumerate}
Neither structure mapping nor neural architectures alone support these five requirements. We argue that HDC can support these requirements and provide concrete guidance for how concepts can be represented and analogized. We call this model a ``conceptual hyperspace.'' CST offers a set of algorithms to solve several types of analogies \cite{VelezGardenforsAnalogyAsSearch}. In this paper, we provide a proof-of-concept for how these algorithms can be implemented with HDC: how concepts can be encoded, how salient relationships can be identified, and how mapping can be performed. We also perform experiments in toy domains to illustrate the approach. 

\color{black}


\section{Background}
\label{sec:background}

In this section, we introduce both the Conceptual Spaces Theory and the Hyperdimensional Computing paradigm, which we will then combine in the next sections.

\subsection{Conceptual Spaces} 
\label{sec:conceptual-spaces} 

The Conceptual Spaces framework \cite{Gardenfors2000,Gardenfors2014} adopts a prototype theory of concept representation \cite{Murphy}, which models concept prototypes as points within a geometric metric space \cite{BellmundSpatialCodes}. This space is constructed from one or more property dimensions of the represented concepts. Properties constitute direct sensory observations or hierarchical abstractions built from sensory observations. One or multiple integral properties constitute a domain. The integral color domain, for example, consists of three property dimensions along the positive Real number line: hue, brightness, and saturation. The weight domain consists of a single property dimension along the positive Real number line. Concepts are convex regions within the space. For concepts that span multiple domains, the domains can be correlated or weighted in various ways.\footnote{We adopt from Conceptual Spaces Theory only the definitions we need to show the efficacy of HDC applied to analogical inference. The interested reader can learn more about CST in \cite{Gardenfors2000,Gardenfors2014}.} 

Figure \ref{fig:colorDomain} illustrates the color domain used as a running example throughout this paper. The prototype for the color concept $\text{\footnotesize{PURPLE}}$, for example, falls at the location $HUE = 315^{\circ}$, $SATURATION = 87$, and $BRIGHTNESS = 53$, within a cylindrical frame of reference. Three other color prototypes and their locations are also shown. The similarity between each of these colors can be modeled by their respective prototype distances from each other. 

\subsection{Hyperdimensional Computing (HDC) }
 \label{sec:hdc}

HDC uses hypervectors for computation. Hypervectors are random high-dimensional (1,000+) vectors that combine hierarchically to produce new hypervectors of the same dimension. The hypervectors entail binary ($\{0, 1\}^d$), bipolar ($\{-1, +1\}^d$), real ($\mathcal{R}^d$), or complex ($\mathbb{C}^d$) samples. A tradeoff typically exists between processing speed and computational power for each sample type. We pick complex samples for our experiments, because they entail all other types and are the most computationally powerful \cite{Plate_2000}. As neuromorphic hardware achieves greater fidelity, however, this tradeoff may change. Researchers have suggested that complex hypervectors map to neural cell assemblies, where the phase of each sample represents the phase of their neuronal spikes \cite{Orchard_SNN}. 

A crucial advantage of high-dimensionality is that the likelihood of two random hypervectors being orthogonal is extremely high. This penchant for orthogonality means hypervectors are capable of encoding scalars and representing bases within a latent representation space. Compositions can be represented in this space without much overlap, while remaining robust to noise. Additionally, through various operations, atomic concepts can be composed symbolically to define new abstract concepts. The HDC community has developed  operations to manipulate these data structures, allowing the creation of a flexible symbolic algebra over the vector space. Several of these operations will be used to compute analogical inference. We represent concepts within a latent space using random complex hypervectors of length equal to $10^4$.

Complex-sampled HDC uniquely affords mechanisms for artificial intelligence such as traditional data structures like trees and graphs \cite{klyko1}, navigation \cite{Komer2020EfficientNU}, probabilistic modeling \cite{FurlongCogArch}, reinforcement learning \cite{QHD_new}, models of Grid and Place cells \cite{Bartlett,Dumont2020AccurateRF}, etc. All of these mechanisms rely on complex-sampled HDC's ability to form kernels out of hypervectors \cite{FradyCompOnFunct}. As we will show in the following sections, we model concepts within conceptual hyperspace as three-dimensional radial basis kernel functions. Section \ref{sec:our-approach} outlines our general approach, and Section \ref{sec:hdc-cst} focuses more on its details. 



\section{Proposed Approach: Conceptual Hyperspaces}
\label{sec:our-approach}

In this section, we begin formally defining the problem setting and our proposed approach for solving the analogical mapping problem. 

\subsection{Problem Definition}
\label{sec:problem}

We return to our compositional analogy of Eq. \ref{eq:analogy}. Here, $\concept{A}$  and $\concept{B}$ are ``source'' concepts and $\concept{C}$ and $\concept{X}$ are ``target'' concepts \cite{Gentner_1998}. The mapping task is to find $\concept{X}$ that satisfies the underlying analogical relationship, namely that the relationship between the source concepts matches the relationship between the target concepts. Consider a domain $\domain$ that is a vector subspace (of, say, $\real{n}$) with $k$ bases. Here, we can initially represent the concepts $\concept{A},\concept{B},\concept{C} \leq \concept{D}$ as themselves being subspaces within $\domain$, and therefore representable within $\domain$. For example, we can think of the ``color'' domain comprising $k=3$ bases -- hue, saturation and brightness -- and the concept of red can be thought of as shades of ``red'' falling within the subspace of color that an agent might consider to be reddish. Because concepts are often vaguely defined, we leverage prototype theory and capture prototypes within concepts, which are meant to represent the concept more precisely. Here, prototypes are individual vectors or points in a domain, such as $\proto{X} \in \domain$. Thus, when computing analogies between concepts   $\concept{A},\concept{B},\concept{C}$, we use prototypes\footnote{We could just as well use non-prototypical points to compute, but we stick to prototypes in this section, because it conforms to our running example within the color domain.} to compute $\proto{A}: \proto{B} :: \proto{C} : \proto{X}$. Since prototypes are vectors in $\domain$, they have projections onto each of the bases of $\domain$, thereby representing the extent to which a prototype extends along that basis. Compositionally, this is a useful notion allowing us to capture how much hue, saturation and brightness the prototypical red has. The $\domain$, together with its bases, allow us to represent concepts as convex regions in $\domain$  and prototypes as points in $\domain$. We can now define an analogical mapping problem in terms of conceptual spaces as follows:

\begin{definition}\label{def:amp}
    \textbf{Analogical Mapping Problem: } 
    Given $\proto{A} \in \concept{A}, \proto{B} \in \concept{B}$ and $\proto{C} \in \concept{C}$, determine $\proto{X} \in \concept{X}$, including projections of $\proto{X}$ along each of $k$ bases of domain $\domain$ such that concepts $\concept{A},\concept{B},\concept{C},\concept{X} \leq D$. 
\end{definition}

An implicit prerequisite in Def. \ref{def:amp} is that the source and target concepts can all be represented within domain $\domain$  using a set of salient $k$ basis, which we will discuss in the next section together with an approach for solving this task.
 
\subsection{Analogical Mapping Algorithm}
\label{sec:algo}

Algorithms 1 to 4 describe the proposed approach to using HDC to solve the analogical mapping problem. Overall, the approach is to encode the prototypes into hyperspace, search for the analogical mapping in hyperspace, and then decode the hypervector to obtain the prototype of the desired target concept. More broadly, our approach has two general steps: (1) ensure saliency requirements are satisfied to construct the computation, and (2) perform the computation of the analogy. 

\begin{algorithm} 
    \caption{Overall}
    \begin{algorithmic}[1] 
    \STATE Input: $\proto{A} \in \concept{A}, \proto{B} \in \concept{B}$ and $\proto{C} \in \concept{C}$
    \STATE Output: $\proto{X} \in \concept{X}$
    \STATE $\hv{a},\hv{b},\hv{c} \gets \mathit{encode}([\proto{A},\proto{B},\proto{C}])$
    \STATE $\hv{x} \gets \mathit{find}([\hv{a},\hv{b},\hv{c}])$
    \STATE $\proto{X} \gets \mathit{decode}(\hv{x})$
    \RETURN $\proto{X}$
    \end{algorithmic}
    \label{algo:overall}
\end{algorithm}

Two assumptions are made in this algorithm:
\begin{enumerate}
    \item That $k$ bases for a domain\footnote{The term ``domain" here refers to a vector subspace, not the CST-specific term.} $\domain$ has been obtained. Such a bases set captures shared properties of concepts $\concept{A},\concept{B},\concept{C}$. In the case of our color example, all the concepts share a common set of three basis. In other property domains\footnote{Here we refer to the CST-specific definition of ``domain."} this may require additional processing, the discussion of which is beyond the scope of this paper. See \cite{Gardenfors2000} for more insight.
    \item That we already have prototypes $\proto{A},\proto{B},\proto{C}$ selected for the concepts. This may require retrieval from long-term memory.
\end{enumerate}

To encode the prototypes, we propose using a Fractional Power Encoding, which allows us to capture gradations along each of our basis in the domain. Algorithm 2 shows us how to encode by first generating basis hypervectors for each of the $k$ dimensions of the domain, $\domain$. These are randomly sampled complex hypervectors from a Gaussian distribution.  To encode, we exponentiate these hypervectors with normalized prototype property values and then bind the $k$ hypervectors together for each prototype. This operation produces a new hypervector of the same $d$ dimensions and serves as a 3D radial basis function kernel in conceptual hyperspace.\footnote{by ``zip'' on line 9, we mean iterator of tuples where they are incremented in pairs in a lock-step manner}  

\begin{algorithm}[h] 
    \caption{encode}
    \begin{algorithmic}[1] 
    \STATE Input: $\proto{A},\proto{B},\proto{C}$
    \STATE Output: $\mathbf{F}$
    \STATE $k \gets \dime{\domain}$
    \STATE $\basis \gets \mathit{sampleBasesHypervectors}(k)$ \label{algo:encode:line4}
    \STATE $ \normal{\proto{A}},\normal{\proto{B}},\normal{\proto{C}} \gets \mathit{normalize}(\proto{A},\proto{B},\proto{C})$ \label{algo:encode:line5}
    \STATE initialize $\mathbf{F}$ to contain the encoded versions of $\normal{\proto{A}},\normal{\proto{B}},\normal{\proto{C}}$ \label{algo:encode:line6}
    \FOR{$\normal{\proto{i}} \in [\normal{\proto{A}},\normal{\proto{B}},\normal{\proto{C}]}$}
    \STATE initialize $\hv{f}_i$
    \FOR{$\rvm,j \in \mathit{zip}(\basis,\normal{\proto{i}})$}
    \STATE $\rvm^* \gets \mathit{exp}(\rvb_i,j)$ \label{algo:encode:line10} \COMMENT {FPE exponentiation} 
    \STATE $\hv{f}_i \gets \hv{f}_i \bind \rvm^*$  \COMMENT {binding operation} \label{algo:encode:line11}
    \ENDFOR
    \STATE $\mathbf{F} \gets \mathit{append}(\hv{f}_i)$
    \ENDFOR
    \RETURN $\mathbf{F}$
    \end{algorithmic}
    \label{algo:encode}
\end{algorithm}
The approach taken to solve an analogical inference problem within conceptual space depends on the type of analogy being calculated. For analogies confined to object categories, CST recommends the Parallelogram model \cite{Rumelhart1973AMF}. We implement this model in hyperspace, as shown in Algorithm \ref{algo:find}, via binding operations with hypervectors. 

\begin{algorithm}[h] 
    \caption{find (using parallelogram method)}
    \begin{algorithmic}[1] 
    \STATE Input: $\hv{a},\hv{b},\hv{c}$
    \STATE Output: $\hv{x}$
    \STATE $\hv{x} \gets (\hv{c} \bind \hv{a}^{-1}) \bind \hv{b}$ \label{algo:find:line3}
    \RETURN $\hv{x}$
    \end{algorithmic}
    \label{algo:find}
\end{algorithm}
$\hv{x}$ represents a latent point in $k$-dimensional conceptual hyperspace, implemented by a $d$-dimensional hypervector in bound superposition. But we don't know the exact location in $k$-space (within domain $\domain$) of the prototype for concept $\concept{X}$, because this information is distributed within the hypervector and not human interpretable. The challenge of ``factorizing'' the components of the hypervector stems from the combinatorial explosion that occurs from explicitly encoding all possible locations in hyperspace and comparing each one against $\hv{x}$. To address this issue, researchers have recently developed so-called resonator networks that enable factorization without such an exhaustive search \cite{Frady2020ResonatorN1}. 

\begin{algorithm}[h] 
    \caption{decode}
    \begin{algorithmic}[1] 
    \STATE Input: $\hv{x}, \mathit{resolution}$
    \STATE Output: $\proto{X}$
    \STATE $k \gets \dime{\domain}$
    \STATE $\mathbf{Q} \gets \mathit{makeCodebook(}\mathit{resolution},k)$
    \STATE $\proto{x} \gets \mathit{resnet}(\hv{x}, \mathbf{Q})$ \label{algo:decode:line5}
    \RETURN $\proto{X}$
    \end{algorithmic}
    \label{algo:decode}
\end{algorithm}

A first step in factorization or decoding is in making a “code book” for each of the k bases of domain $\domain$. A code book, $\mathbf{Q}$, is a set of reference hypervectors built from each basis in the domain. If the domain is $\real{k}$, then there are $k$ codebooks. We discretize each dimension into discrete points, depending on our desired resolution. For example, when handling colors, we can normalize each of the $k$ dimensions into a range of [-10,10] and then assign one point for each 0.5 increment, creating 41 locations along each of our 3 basis dimensions of the color domain. If we assign a hypervector to each of these locations, we obtain three code books, each containing $41$ $d$-dimensional hypervectors. We then employ a resonator network to iteratively discover the underlying factors for $\proto{X}$.

\section{The Conceptual Hyperspace} 

\label{sec:hdc-cst}


\subsection{Encoding Concepts in Hypervectors}
\label{sec:encoding}
To detail our approach using HDC for encoding conceptual hyperspace, we return to our running example of a composed analogy, this time associated with colors. 
$$\concept{PURPLE} : \concept{BLUE} :: \concept{ORANGE} : \concept{X}$$

\begin{figure}
    \centering
    \begin{minipage}{0.45\textwidth}
        \centering
        \includegraphics[width=.9\textwidth]{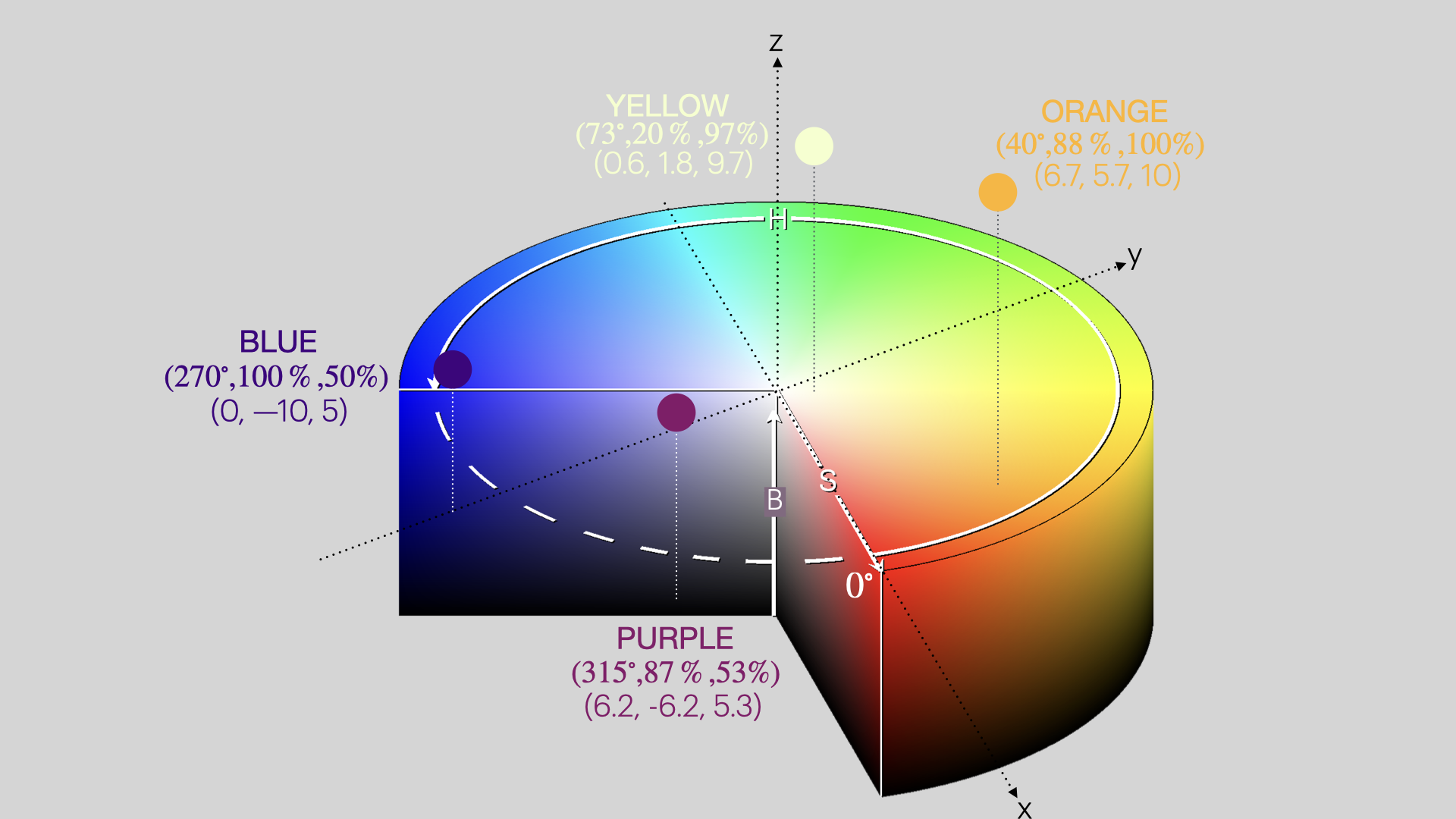} 
        \caption{Color domain showing the location of prototype points for PURPLE, BLUE, ORANGE, and YELLOW.}
        \label{fig:colorDomain}
    \end{minipage}
\end{figure}

Figure \ref{fig:colorDomain} shows the geometry of this composed analogy within the color domain. The example starts by building three three-dimensional concept regions (for the operands purple, blue, and orange) within the color domain. Since the color domain is a three-dimensional space, we begin by initializing three \textit{basis} hypervectors, which define the space (Algorithm \ref{algo:encode}, Line \ref{algo:encode:line4}).

\begin{figure}
    \centering
    \begin{minipage}{0.45\textwidth}
        \centering
        \includegraphics[width=0.9\textwidth]{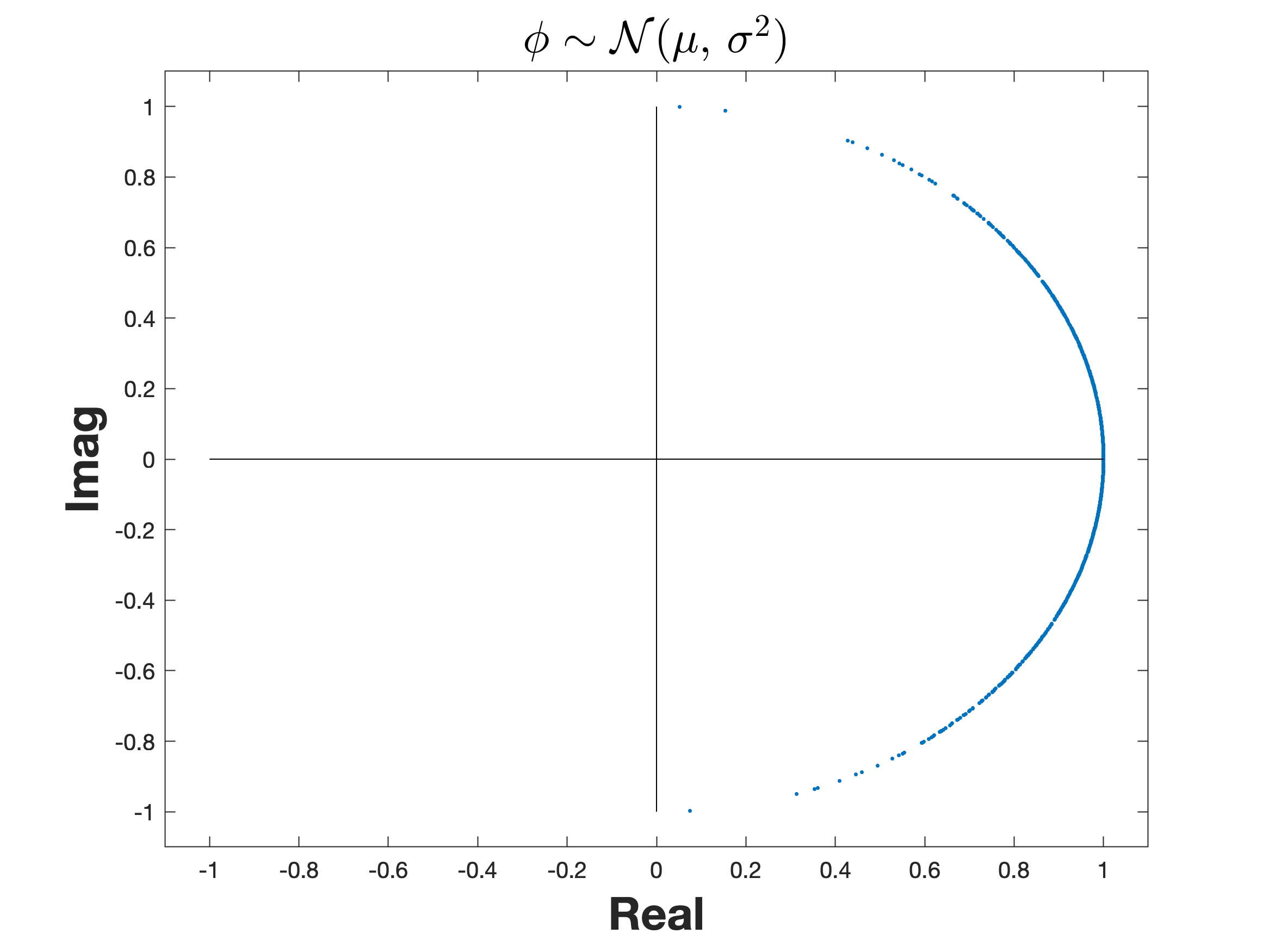} 
        \caption{A complex-sampled hypervector of dimension $d=1000$ samples initialized to a Gaussian phase distribution around the unit circle: $\phi \sim \mathcal{N}(\mu,\,\sigma^{2})$, where here $\mu = 0$ and $\sigma = \frac{\pi}{7}$. We experimented with various $\sigma$ values.}
        \label{fig:gaussianDist}
    \end{minipage}
\end{figure}

A hypervector, $\hv{x}$, in conceptual hyperspace initializes to a Gaussian phase distribution around the unit circle: 
$$ \hv{x} \in \mathbb{C}^d \text{, where sample } x_j = e^{i\phi_j}, $$ 
with phases $\phi_j \sim \mathcal{N}(\mu,\,\sigma^{2})$, where $\mu$ is the mean phase and $\sigma$ the standard deviation in radians. See Figure \ref{fig:gaussianDist}. Bases hypervectors with large enough $\sigma$ initialize to orthogonal. 

The agent must encode the color properties into a concept space within its working memory. A visual preprocessing step extracts a concept's property values from either a specific object in the environment or from prototypes in long-term memory (steps beyond the scope of this work). For the rest of this paper, we will presume to use prototypes rather than observations. The three color property values then get encoded into three hypervectors, one for each basis, which represent prototype locations along their respective dimensions (for example, encoding the amount of hue, saturation, and brightness for each color). This encoding is done via a process called $\textit{Fractional Power Encoding}$ (Algorithm \ref{algo:encode}, Line\ref{algo:encode:line10}) \cite{Komer2019ANR}. The prototype locations may also get stored into temporary variables to be used later. Both FPE and variable storage apply one of HDC's most fundamental operations, called $\textit{binding}$. 

\begin{figure}
    \centering
    \begin{minipage}{0.35\textwidth}
        \centering
       \includegraphics[scale=0.35]{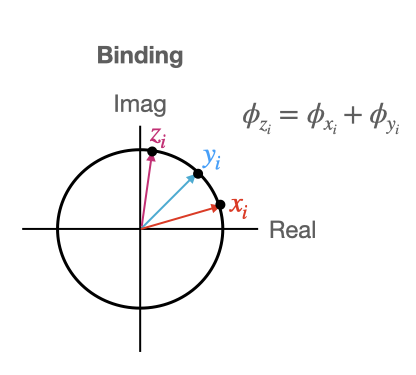} 
        \caption{Result of one sample-wise binding operation between two complex hypervectors.}
        \label{fig:binding}
    \end{minipage}
\end{figure}

The binding process is simultaneously a symbolic variable operation and a signal processing operation. Binding two HDC \footnote{ Throughout this paper we simply refer to complex-sampled hypervectors as ``HDC" hypervectors. A synonymous name for hypervectors with complex sample type is Fourier Holographic Reduced Representation (FHRR).} hypervectors performs a circular convolution in Fourier space, represented by the operator $\circledast$, which is a sample-wise multiplication of the two hypervectors. Given two hypervectors $\hv{x} \in \mathbb{C}^d \text{ and } \hv{y} \in \mathbb{C}^d$, the result of their binding is
\begin{equation}\label{eqn:binding}
    \hv{z} = \hv{x} \circledast  \hv{y}.
\end{equation}
Figure \ref{fig:binding} illustrates how binding adds the phases of each respective pair of samples, given by $\phi_{z_i} = \phi_{x_i} + \phi_{y_i}$. $\hv{z}$ results in a hypervector orthogonal to both $\hv{x}$ and $\hv{y}$ if $\hv{x}$ and $\hv{y}$ start orthogonal to each other. 

\begin{figure}[h]
    \centering
    \begin{minipage}{0.35\textwidth}
        \centering
        \centering
        \includegraphics[scale=0.35]{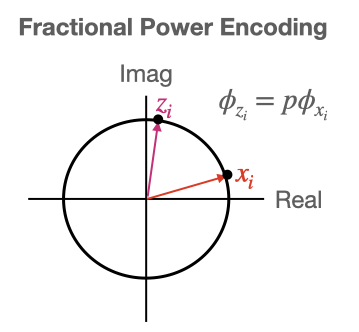} 
        \caption{Result of fractional power encoding for one sample within a complex hypervector.}
        \label{fig:FPE}
    \end{minipage}\hfill
    
\end{figure}

Unbinding is the inverse of binding and is performed by binding with the complex-conjugate of one operand. For example, given Equation \ref{eqn:binding},
\begin{align*}
 \hv{x} = \hv{z} \circledast  \hv{y}^{-1},
\end{align*}    
 where $\hv{y}^{-1}$ is the complex-conjugate of $\hv{y}$, making the phase relationship between samples $\phi_{x_i} = \phi_{z_i} - \phi_{y_i}$. Unbinding is used to dereference a value from a variable.
 Binding is the base operation for FPE. FPE encodes a scalar into a hypervector. Given a hypervector, $\hv{x} \in \mathbb{C}^d$, and prototype location scalar, $p \in \mathbb{R}$, 
$\hv{z}$ encodes $p$ via exponentiation:
\begin{equation}\label{eqn:FPEBackground2}
    \hv{z} = \hv{x}^{p} = \hv{x} \circledast  \hv{x} \circledast  ... \circledast \hv{x} \text{ (p times)}.
\end{equation}
In Equation \ref{eqn:FPEBackground2}, the hypervector $\hv{x}$ appears $p$ times. $p$ can be any Real value \cite{Komer2020EfficientNU}. Figure \ref{fig:FPE} shows what happens to the phase of each sample of $\hv{x}$ after an FPE encoding. The initial phase gets multiplied by $p$, and the encoding process behaves as $p$ sequential bindings, even if $p$ is not an integer. 

By binding together our three fractionally power encoded basis hypervectors (Algorithm \ref{algo:encode}, Line\ref{algo:encode:line11}) , we build a latent concept prototype within the color domain of conceptual hyperspace. For example, the final concept encoding for $\concept{PURPLE}$ looks like this:
\begin{equation} \label{prototypeLocation}
    \concept{PURPLE} = \hv{x}^{6.2} \circledast \hv{y}^{-6.2} \circledast \hv{z}^{5.3},
\end{equation}
where we establish Cartesian locations $p_1 = 6.2$, $p_2 = -6.2$, and $p_3 = 5.3$ for the ``prototype location" based on the following pre-processing steps (Algorithm \ref{algo:encode}, Line \ref{algo:encode:line5}), where $\beta = 10$ is a scaling constant, and we use the values $HUE = 315^{\circ}$, $SATURATION = 87$, and $BRIGHTNESS = 53$:
\begin{align*}
    & p_1=cos(HUE)\cdot SATURATION \slash \beta, \\
    & p_2=sin(HUE)\cdot SATURATION \slash \beta, \\
    & p_3= BRIGHTNESS \slash \beta.
\end{align*}

Because we initialized each basis to a Gaussian distribution, our concept in hyperspace, shown in Equation \ref{prototypeLocation}, retains the computational properties of a latent three-dimensional radial basis kernel embedded within a single d-dimensional complex-sampled hypervector. 

\subsection{Analogical Mapping Algorithm for Category-based analogies}
\label{sec:algorithm}

Researchers have classified the semantic relationships present in analogy (e.g., the colon in $A:B$) into many different types \cite{Collins_Burstein_1987,VelezGardenforsAnalogyAsSearch} that can broadly fall into ``category-based'' and ``property-based.''  There are other classifications like event-based, part-whole, and causal; but for this paper we focus on just the first two.

Recall our example from above. 
$$\concept{PURPLE}:\concept{BLUE}::\concept{ORANGE}:\concept{X}$$ 
We solve for $\concept{X}$, after first guaranteeing $\concept{A}$, $\concept{B}$, and $\concept{C}$ satisfy the algorithm-specific pre-processing requirements. For this algorithm, there are two pre-processing requirements:
\begin{enumerate}
    \item Check that concepts $\text{\footnotesize{A}}$, $\text{\footnotesize{B}}$, and, $\text{\footnotesize{C}}$ are labeled in long-term memory as a common superordinate category.  Labelling and label checking can be accomplished through a variety of methods, including Kanerva's distributed record \cite{Kanerva1997FullyDR}, graphs, tree structures, or state machines,  \cite{klyko1}. Each of these symbolic data structures can be built with HDC and  efficiently store properties of prototypes within long-term memory or perform other symbolic logic. The detailed implementation of this long-term memory is beyond the scope of this paper and a work in progress. 
    \item Find the dimensions over which each $\text{\footnotesize{A}}$ and $\text{\footnotesize{B}}$ have common properties. This step can also be solved using HDC methods and is beyond the scope of this paper. 
    \end{enumerate}
 In this example, all operands are already within the color category (which also happens to be a single domain), so no algorithm-specific pre-processing needs to happen. 

Since all concepts are now known to be within the same category, it is straightforward to find the location for the prototype of $\concept{X}$. We apply the Parallelogram model (Algorithm \ref{algo:find}, Line \ref{algo:find:line3}) \cite{Rumelhart1973AMF}: 
\begin{equation} \label{eq:parallelogram}
    \proto{X} = \proto{C} - \proto{A} + \proto{B}
\end{equation}
Encoding these locations into hypervectors by applying Equation \ref{eq:parallelogram} within a conceptual hyperspace and applying the appropriate HDC operations, gives us
\begin{equation}\label{eq:parallelogramHDC}
    \hv{x} = (\hv{o} \bind \hv{p}^{-1}) \bind \hv{b},
\end{equation}
where we have abbreviated the colors to the first letter of their names. It's worth noting a few things at this point. We know from the statistical literature that the convolution of two Gaussian distributions adds their means \cite{jaynes03}; and because each complex hypervector lives within the Fourier domain, the sample-wise multiplication of the binding operator performs a convolution. All elements of the operand hypervectors compute within bound superposition of three encoded basis per concept. The result hypervector, $\hv{x}$, remains entangled within bound superposition and remains random. Nevertheless, the phase changes induced by the algebraic operations used to compute the analogy maintain their statistical fidelity within the answer. 

While we have access to the hypervector elements that comprise our answer's prototype, we do not yet know its encoded latent Cartesian location $(p_1, p_2, p_3)$. This location is deeply encoded into the aforementioned bound superposition. To find an estimate of the encoded location within each basis hypervector, we employ an HDC resonator network \cite{Frady2020ResonatorN1} whose job is to find factors like these within bound superposition. Most mathematical details about how resonator networks work are beyond the scope of this paper. What is important, however, is this: If we build a code book that contains all possible encoded hypervector factors as a reference, then the resonator network can quickly factor the bound superposition. The network can actually take advantage of statistical properties afforded by this bound superposition rather than get hindered by them. The number of iterations needed to find factors scales linearly $\mathcal{O}(M)$, where $M$ is the search space size. Kent's PhD thesis \cite{KentThesis} shows how resonator networks factor bound superposition problems faster than any other known method. 

The resonator network ``collapses" the bound superposition from a learned or innate range of typical property values, resulting in an estimate for the prototype location, $\proto{X}$, when it returns references to the set of factors. In this experiment, our resonator network took two iterations to find the correct encoded basis factors for an estimate of $\proto{YELLOW}$ (Algorithm \ref{algo:decode}, Line \ref{algo:decode:line5}). Our model's code book included 41 different possible encoded basis hypervectors for each dimension, spanning each basis' range of possible normalized values from $-10$ to $+10$, at increments of $0.5$. In our experiment, the resonator network returned the location $p_1 = 0.5$, $p_2 = 2.0$, and $p_3 = 9.5$, which equates to un-normalized values of $HUE=75^{\circ}$, $SATURATION=75$, and $BRIGHTNESS=95$. This is as close as it could theoretically approach the prototype location for $\text{\footnotesize{YELLOW}}$, given the $0.5$ normalized resolution of our code books.

\subsubsection{Solving a Property-based Analogy}
This type of analogy compares a semantic relationship between categories and properties. Recall the general representation for analogy:
\begin{equation*}
    \concept{A} : \concept{B} :: \concept{C} : \concept{X}
\end{equation*}
A property-based example provided by \cite{VelezGardenforsAnalogyAsSearch} is 
\begin{equation*}
    \concept{APPLE}:\concept{RED}::\concept{BANANA}:\concept{X}.
\end{equation*}
The steps to solving this composed analogy are the following: 
\begin{enumerate}
    \item Identify the salient property dimensions that correspond to concept or property $\concept{B}$. In this example, this would be the bases dimensions associated with $\concept{RED}$ (i.e. the color domain).
    \item Find the location within the identified salient dimensions closest to prototype location $\concept{C}$. In this example, this would be the location along the hue, saturation, and brightness bases associated with concept $\concept{BANANA}$ (i.e. $\concept{YELLOW}$).
\end{enumerate}    
In general, the agent will often need to search for salient properties. This search requires a distance metric. For example, in this problem the agent needs to search for what bases constitute $\concept{RED}$. The agent stores bases in long-term memory, because these bases hypervectors are needed to build code books. To determine the salient bases associated with $\concept{RED}$, the agent computes a distance metric between all domain bases hypervectors stored in memory and the domain basis hypervector used to encode $\concept{RED}$. The closest domain hypervector in memory to $\concept{RED}$'s domain hypervector is the salient one. By using the properties of HDC Gaussian kernels \cite{FradyCompOnFunct}, a simple dot-product between hypervectors measures their relative distance from each other within conceptual hyperspace. \footnote{Each concept or basis hypervector is also a Gaussian kernel within a formal Reproducing Kernel Hilbert Space (RKHS). This fact affords the dot-product as a distance metric.} This brings us to the final part of our introduction to HDC. Similarity between complex-sampled unitary hypervectors is defined as the mean of the cosine of angle differences between corresponding samples \cite{Vine2010SemanticOE}. This definition equates to the inner product between two complex hypervectors. Given $\hv{x} \in \mathbb{C}^d, \hv{y} \in \mathbb{C}^d$, the relative distance between $\hv{x} \text{ and } \hv{y}$ is  $similarity ( \hv{y},\hv{x} )=\hv{y}^{-1} \cdot \hv{x}$. There could be myriad types of search depending on analogy type. This example shows only one type of search. But given the algebraic, symbolic, and metric operations afforded by HDC, many different types of search are possible.

\section{Discussion}

\subsubsection{The Importance of HDC as a Modeling Tool}
 Why should we care about using a neurally-plausible analogy engine? Why can't we just perform our analogical number crunching with a base-10 number system? The traditional approach would certainly be more straightforward. 

The reason is because the more traditional approaches are stuck at Marr's algorithmic level \cite{Marr}. Constraining our representations to be neurally-plausible adds scientific value to our model. If we can achieve analogical inference by using a model between Marr's algorithmic and implementation levels, which we propose here, then we've reduced the search space for an algorithm-plus-implementation towards human-level intelligence. 

At the engineering level, we concede that for the toy model presented here, a neurally-plausible conceptual hyperspace seems more complex than a traditional conceptual space using traditional vectors. But the brain does not use a von Neumann architecture. As we scale this model to include more cognitive functionality that would require more resources, we could potentially build it within a spiking neural network or other power-efficient neuromorphic hardware. Constraining our AI models to neural-plausibility affords hope for human-level cognitive efficacy at scale. 

\subsubsection{The Origin of Property Dimensions}
The origin of property dimensions do not yet seem to be deeply grounded in theory. Therefore, we plan to pursue the best ways to model these. In this paper, we treat property dimensions as orthogonal bases within a conceptual hyperspace, which act as building blocks for intrinsic domains. This conveniently works out from both a signal processing and cognitive science perspective. The signal processing theory literally requires orthogonality to afford kernel construction. If it turned out that the brain built a hierarchy of property dimensions from a finite set of orthogonal basis dimensions then this would also be elegantly satisfying for cognitive science. \cite{Wierzbicka} provides a finite set of semantic primitives over all languages, which seems like a good place to start building such a conceptual hyperspace model grounded on a finite set of atomic property dimensions. Through HDC operations, we could then generate hierarchical property dimensions on the fly that correspond with analogy algorithm requirements. 

Neuroscience experiments that take place in fMRI machines show evidence that the Entorhinal cortex-hippocampus system quickly builds highly specific hierarchical concept regions. For example, the regions reported in \cite{BellmundSpatialCodes} are ``Neck length" versus ``Leg length." We can possibly use HDC to teach an agent how to build the appropriate property dimensions. HexSSP, which are HDC kernel hypervectors that model Place and Grid cells, introduced by \cite{Bartlett}, may afford learnable resolution sizes for property dimensions. We would like to integrate this learning of property dimensions with efficient resonator network code book design, as well.

\subsubsection{Exploration of Kernel Functions}
Playing with the bandwidth on our radial basis kernel functions allows us to shape the similarity regions within conceptual hyperspace. Given the flexibility of HDC, radial basis functions are not the only kernel function at our disposal, however. Playing with different kernel types and their respective parameters afford myriad concept region shapes \cite{FradyCompOnFunct} for a variety of semantic similarity. Most kernel similarities adhere to the (arguably) soft CST requirement of maintaining domain convexity \cite{Hernandez-Conde2017}. But if we'd like to model non-convexity, then that's possible too. It's possible (albeit inelegant) to symbolically label any concept by binding to it an additional hypervector, which can be stripped off before signal processing begins. \cite{Balkenius2016} discuss radial basis function network models for learning within the context of motor movement, reasoning, and other applications. The agent can potentially learn the appropriate kernels to use along with their respective parameters. 

\subsubsection{Conceptual Hyperspace as a Generative Model}
HDC allows us to build novel concept regions in a generative manner. If the answer to an analogy problem generates a concept location that does not, say, already have a linguistic or symbolic label within a minimal distance of an existing prototype, then the agent has an opportunity to be creative. The way we've modeled a concept space with kernel functions does not require the entire space to be tiled with concepts. Any time a new concept is produced, the agent can use the kernel properties of HDC to quickly find the new concept's distance to all existing prototypes and decide what it wants to do -- create a new prototype, merge with an existing concept, implementing a sort of exemplar model via additive HDC superposition (an HDC operation not covered in this paper), or do nothing, allowing the agent to retain a more holistic conceptualization. In this manner, HDC has the potential to extend CST into a generative framework not constrained to a rigid theory of prototypes. 

We know that analogy is the ``Fuel and Fire of Thinking" as Douglas Hofstadter \cite{Hofstadter} likes to say. Therefore we know that analogical inference likely plays a role in all forms of cognition. This is a core principle that will guide us as we move forward with this research.

\appendix

\bibliographystyle{named}
\bibliography{ijcai24}

\end{document}